\documentclass[runningheads]{llncs}

\usepackage[T1]{fontenc}
\usepackage{graphicx}
\usepackage{booktabs}
\usepackage{multirow}
\usepackage{amsmath,amssymb}
\usepackage{siunitx}
\usepackage{url}
\usepackage{adjustbox}
\usepackage{float}

\newcommand{\equalcontrib}
{\textsuperscript{\ensuremath{\dagger}}}
\newcommand{\corrauth}{\textsuperscript{*}}
\usepackage{cite}
\begin{document}

\title{Conditional Latent Diffusion Model with Fourier-based Motion Modelling for Virtual Population Synthesis}
\titlerunning{4D F-MeshLDM}

\author{Shaokun Lan\inst{1,2,3}\equalcontrib \and
Haoran Dou\inst{1,2,3}\equalcontrib \and
Jinghan Huang\inst{1,2,3}\and 
Arezoo Zakeri\inst{1,2,4}\and
Fengming Lin\inst{1,2,3}\and
Zherui Zhou\inst{1,2,5}\and
Jinming Duan\inst{1,2,4}\and
Alejandro F. Frangi\inst{1,2,3,4,6}\corrauth}


\authorrunning{S. Lan et al.}

\institute{
Centre for Computational Imaging and Modelling in Medicine (CIMIM), University of Manchester, Manchester, UK\\
\and
Christabel Pankhurst Institute, University of Manchester, Manchester, UK\\
\and
Department of Computer Science, University of Manchester, Manchester, UK\\
\and
Division of Informatics, Imaging \& Data Sciences, University of Manchester,
Manchester, UK\\
\and
Department of Electrical \& Electronic Engineering, University of Manchester, Manchester, UK\\
\and
NIHR Manchester Biomedical Research Centre, Manchester Academic Health Sciences Centre, University of Manchester, Manchester, UK\\
\email{alejandro.frangi@manchester.ac.uk}
}

\maketitle

\begingroup
\renewcommand{\thefootnote}{%
  \ifcase\value{footnote}%
  \or *%
  \or \ensuremath{\dagger}%
  \else\arabic{footnote}%
  \fi}
\footnotetext[1]{\raggedright Corresponding author: Alejandro~F.~Frangi.\\
Email: \url{alejandro.frangi@manchester.ac.uk}\par}
\footnotetext[2]{\raggedright Shaokun Lan and Haoran Dou contributed equally to this work.\par}
\endgroup

\begin{abstract}
\textit{In-silico} trials of medical devices require the generation of virtual populations of anatomies. In cardiovascular applications, virtual anatomy is typically represented as a 3D+t mesh sampled from a generative model. However, most existing mesh generators focus on static anatomy, while sequence models often lack explicit periodicity. To this end, we propose 4D F-MeshLDM, a conditional generative framework comprising a convolutional mesh VAE to encode meshes, a structural latent space that parameterises motion using a truncated Fourier series, and a diffusion prior that learns the latent distribution over Fourier coefficient tokens. By conditioning the diffusion process on clinical covariates via affine modulation, we enable controllable synthesis. Sampling tokens and performing inverse Fourier synthesis yield cycle-consistent latent trajectories, which can be decoded into 3D+t cardiac mesh sequences. Experiments on 5{,}000 UK Biobank subjects demonstrate that 4D F-MeshLDM outperforms state-of-the-art baselines in anatomical fidelity and achieves near-zero cycle closure error. Furthermore, the generated cohorts accurately preserve clinical functional indices, highlighting the potential of our framework for reliable \textit{in-silico} cardiac trials.

\end{abstract}

\section{Introduction}

As an alternative to clinical trials, \textit{in-silico} trials use computational models to evaluate the safety and efficacy of medical devices \cite{konduri2020silico,pappalardo2019silico}. \textit{In-silico} trials require a diverse virtual population that captures the anatomical and physiological variability of real patients, enabling device performance to be assessed across different cases. In cardiac applications, the virtual population can typically be represented as a generative model that produces realistic anatomy with plausible dynamic motion across the full cardiac cycle.

Statistical shape models and deep generators, like PCA and VAE-based mesh models, have been widely used to model cardiac anatomy \cite{dou2023conditional,dou2024generative,gooya2015bayesian,mozyrska20253d}, but most methods generate static shapes. To address this issue, several deep generative approaches have been proposed for spatio-temporal synthesis of cardiac anatomy \cite{cheart2024,qiao2025personalized}, including implicit spatio-temporal fields \cite{2024stndf}, continuous-time generative dynamics \cite{ma2025cardiacflow}, and disentangled population motion model \cite{dou2025cardiosynth}. However, these methods encourage periodicity implicitly through temporal conditioning, which can lead to drift and imperfect cycle closure, which are undesirable for device evaluation and simulation.

In this work, we propose 4D F-MeshLDM, a conditional latent diffusion model for full-cycle cardiac mesh generation that enforces cycle periodicity and consistency by construction. The cardiac motion of each subject is encoded as a compact set of Fourier coefficient tokens via a truncated Fourier series in the latent space of a mesh VAE, and a transformer-based diffusion prior is trained to model the distribution of these tokens conditioned on clinical covariates. The main contributions are as follows:

\begin{itemize}
    \item We propose the first cardiac motion generative model that guarantees mathematically exact cycle consistency by construction. By parameterising latent trajectories as truncated Fourier series, we eliminate endpoint drift and closure errors, strictly enforcing periodicity without relying on learned recurrent dynamics\cite{cheart2024} or soft temporal conditioning\cite{ma2025cardiacflow, dou2025cardiosynth, 2024stndf}.
    \item To the best of our knowledge, we introduce the first conditional latent diffusion model for 3D+t mesh sequences. We model the complex joint distribution of motion and anatomy in a compact spectral domain via a Transformer-based diffusion prior operating on Fourier coefficient tokens, enabling scalable and controllable synthesis.
    \item Experiments demonstrate that 4D F-MeshLDM significantly outperforms state-of-the-art baselines \cite{cheart2024,ma2025cardiacflow,dou2025cardiosynth}. We achieve nearly zero cycle closure error and superior anatomical fidelity, validating the model's reliability for \textit{in-silico} trials.
\end{itemize}

\section{Methodology}
\begin{figure}[t]
    \centering 
    \includegraphics[width=1\linewidth]{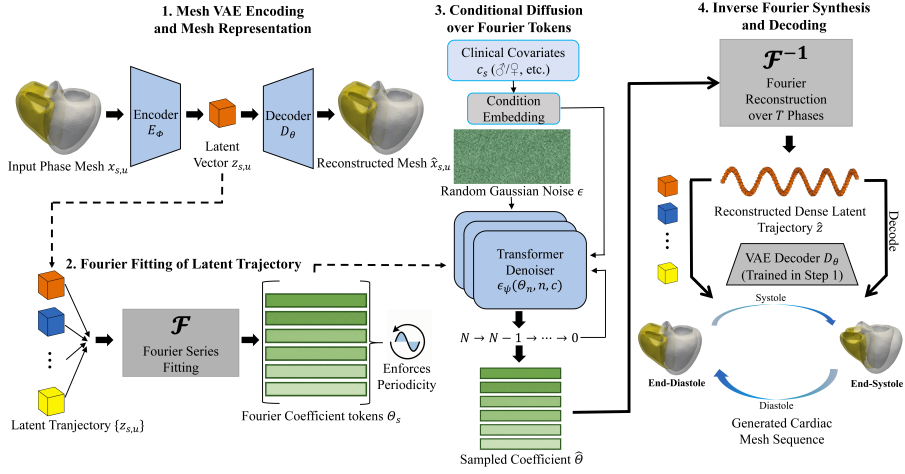}
    \caption{Overview of the 4D F-MeshLDM pipeline. Dashed arrows denote training-time encoding and fitting, while solid arrows denote the generative pathway.}
    \label{fig:pipeline}
\end{figure}

Our framework aims to generate anatomically plausible and cycle-consistent 3D+t cardiac mesh sequences conditioned on clinical covariates. As illustrated in Fig.~\ref{fig:pipeline}, our framework decouples the spatial representation of cardiac anatomy from the modeling of its periodic motion trajectory. A mesh VAE encodes the mesh from each phase to a latent code. Then, cardiac motion is represented by fitting a truncated Fourier series to the latent trajectory, producing a set of coefficient tokens. A conditional diffusion prior models the token distribution, followed by inverse Fourier synthesis with VAE decoding, which generates a full cardiac cycle.

\subsection{Anatomical Latent Representation}
We train a mesh variational autoencoder based on CoMA \cite{ranjan2018coma} to learn compact spatial features. For a mesh $x_{s,u}$ of a given subject $s$ at frame $u$, the encoder $E_\phi$ maps geometry to a latent distribution $q_\phi(z_{s,u}\mid x_{s,u})$. The decoder $D_\theta$ reconstructs the mesh $\hat{x}_{s,u}$ using Chebyshev spectral convolutions and hierarchical pooling. We optimize the standard VAE objective with a weight $\beta$:
\begin{equation}
\mathcal{L}_{\mathrm{VAE}} = \mathbb{E}_{q_\phi}\left[\| x_{s,u} - \hat{x}_{s,u} \|_2^2\right] + \beta\, \mathrm{KL}\big(q_\phi(z_{s,u} \mid x_{s,u}) \,\|\, \mathcal{N}(0, I)\big).
\end{equation}

After training, the encoder mean $z_{s,u} \in \mathbb{R}^D$ serves as the deterministic representation for motion modelling.

\subsection{Cardiac Motion Modelling via Fourier Fitting}
Our framework models cardiac motion in the frequency domain to separate the spatial trajectory of the motion from its cardiac cycle duration. For a subject $s$ with a sequence length of $T_s$ frames indexed by $u=0, \dots, T_s-1$, we define the normalised phase at frame $u$ as $\tau_u = \frac{u}{T_s - 1} \in [0, 1]$. The latent trajectory is approximated by a truncated Fourier series of order $K$:
\begin{equation}
z_{s,u} \approx a_{s,0} + \sum_{k=1}^{K-1} \left( a_{s,k} \cos(2\pi k \tau_u) + b_{s,k} \sin(2\pi k \tau_u) \right),
\label{eq:fourier}
\end{equation}
where $a_{s,k}, b_{s,k} \in \mathbb{R}^D$ are subject-specific coefficients. Crucially, this phase definition implies $\tau_0=0$ and $\tau_{T_s-1}=1$. Since the Fourier basis is 1-periodic, this formulation guarantees mathematically exact cycle closure, ensuring $z_{s,0} \equiv z_{s, T_s-1}$ by construction.

We can express Eq.~\eqref{eq:fourier} linearly. We define a basis vector $A_u \in \mathbb{R}^{1 \times B}$ ($B=2K-1$) containing the harmonic terms:
\begin{equation}
A_u = \big[1,\cos(2\pi \tau_u),\sin(2\pi \tau_u),\dots,\cos(2\pi (K-1) \tau_u),\sin(2\pi (K-1) \tau_u)\big].
\end{equation}

The coefficients are stacked into a matrix $\Theta_s \in \mathbb{R}^{B \times D}$, where rows correspond to the terms $\{a_{s,0}, a_{s,1}, b_{s,1}, \dots\}$. The trajectory is thus given by $z_{s,u} \approx A_u \Theta_s$.

To obtain the ground-truth coefficients $\Theta_s$ for training, we stack the observed frame-wise latents $\{z_{s,u}\}_{u=0}^{T_s-1}$ into a trajectory matrix $Z_s \in \mathbb{R}^{T_s \times D}$. We then solve the following least-squares equation:
\begin{equation}
\Theta_s = \arg\min_{\Theta} \| {A}_s \Theta - Z_s \|_F^2,
\label{eq:ls}
\end{equation}
where ${A}_s \in \mathbb{R}^{T_s \times B}$ is the subject-specific basis matrix constructed using the subject's cycle length $T_s$. Crucially, while ${A}_s$ depends on the temporal sampling $T_s$, the resulting coefficients $\Theta_s$ represent the time-invariant motion feature of the subject. In the following discussion, we omit the subscript $s$ and denote $\Theta$ as the token set for brevity.

\subsection{Conditional Diffusion on Fourier Tokens}
We treat the coefficient matrix $\Theta$ as a sequence of $B$ tokens and model their joint distribution $p(\Theta \mid c)$ using a Denoising Diffusion Probabilistic Model (DDPM) \cite{ho2020ddpm}. The forward process diffuses the coefficients into Gaussian noise over $N$ steps. The reverse generative process is learned by a denoiser network $\epsilon_\psi$, implemented as a Transformer.

The Transformer treats the $B$ frequency components as a sequence of tokens. To preserve frequency information, we augment each token with a learnable Fourier-index embedding. Clinical covariates $c$ are embedded and injected into the network via Adaptive Layer Normalisation (AdaLN) \cite{peebles2023scalable}, allowing the model to generate motion patterns specific to the target population. The objective is to predict the noise $\epsilon$ added at step $n$:
\begin{equation}
\mathcal{L}_{\mathrm{Diff}} = \mathbb{E}_{\Theta_0, \epsilon, n, c} \left[ \| \epsilon - \epsilon_\psi(\Theta_n, n, c) \|_2^2 \right].
\end{equation}

\subsection{Generative Inference}
The generative inference process is detailed below. First, a coefficient matrix $\hat{\Theta} \sim p_\psi(\Theta \mid c)$ is sampled from the conditional diffusion prior on covariate $c$. This determines the time-invariant geometric motion pattern of the heart.  Second, we dynamically adjust the cycle duration to reflect the physiological relationship between demographics and heart rate. We model the duration of the target cycle $T'$ as a function of covariate $c$ with $T' = f(c)$, where $f(\cdot)$ is a mapping derived from the statistics of the training population. Unlike existing methods that require additional data processing or ad-hoc temporal resampling to alter sequence lengths, our method inherently supports generating populations with variable cardiac cycle durations. Then, based on this subject-specific duration $T'$, a dynamic sampling basis matrix $\mathbf{A}' \in \mathbb{R}^{T' \times B}$ can be constructed using the normalised phases $\tau' = \{0, \frac{1}{T'-1}, \frac{2}{T'-1} \dots, 1\}$. The dense latent trajectory is then reconstructed by combining the generated morphology with the subject-specific temporal basis:
\begin{equation}
\hat{Z} = \mathbf{A}' \hat{\Theta}.
\end{equation}

Finally, the full mesh sequence is obtained by decoding each frame: $\hat{x}_u = D_\theta(\hat{z}_u)$. Our work yields a strictly periodic mesh sequence that is consistent with both the morphological and temporal characteristics of the target population defined by $c$.

\section{Experimental Results}
\subsection{Dataset}
We conducted evaluation experiments on a cohort of 5{,}000 time-resolved biventricular (BiV) surface meshes reconstructed from cine CMR available from the UK Biobank (UKBB). For perfectly fair comparisons with baselines which are constrained to fixed temporal lengths, we specifically focused on subjects where all $T_s = T = 50$ in this work. For these subjects, $T=50$ covers the full cardiac cycle starting from end-diastole (ED) at $u=0$, with $u=49$ being the last frame returning to ED. All meshes share a common BiV template topology and consistent vertex ordering across subjects and cardiac phases after reconstruction, enabling point-to-point geometric evaluation. Each mesh was reconstructed by \cite{xia2022automatic}. The cohort was split into 4{,}000/500/500 for training, validation, and testing, respectively. Gender and age were selected as the covariates in this work. We synthesised a virtual cohort with the same number of subjects in the test set to ensure a fair comparison.

\subsection{Implementation Details}
All experiments were implemented in PyTorch and trained on 1 NVIDIA H200 GPU with a total memory of 141 GB. We trained a mesh VAE with latent dimension $D=64$ using AdamW with learning rate \(5\times10^{-4}\) and KL weight $\beta=10^{-5}$ for 400 epochs. To model cardiac dynamics, we fitted a truncated Fourier series with a frequency $K=13$, which achieved the best performance amongst all $K$ values in the ablation study. A tokenized Transformer DDPM (width 256, depth 8, 8 heads) using 500 diffusion steps was trained with learning rate \(3\times10^{-4}\) for 1{,}000 epochs.

\subsection{Baselines and Evaluation Metrics}
The 4D F-MeshLDM was compared with three state-of-the-art cardiac motion generation baselines: (1) CHeart~\cite{cheart2024}, a spatio-temporal mesh generator based on recurrent neural networks; (2) 4D CardioSynth~\cite{dou2025cardiosynth}, a VAE-based synthesis method with spatio-temporal disentanglement; and (3) CardiacFlow~\cite{ma2025cardiacflow}, a continuous-time generative dynamics model. 

We evaluated the distribution quality of anatomy generation using specificity \cite{davies2009building}, measured as the nearest-neighbour distance from generated samples to the real population, and coverage@5 \cite{naeem2020reliable}, measured as the fraction of real samples that have at least one generated sample within the 5-nearest-neighbourhood. To evaluate the spatio-temporal fidelity of subject-specific reconstructions, we calculated the Sequence Root Mean Square (RMS) error, which is defined as the root mean square of the Euclidean distances between corresponding vertices of the generated and ground-truth meshes, averaged across all time frames within the sequence. Moreover, following prior work on spatiotemporal regularisation in cardiac motion analysis \cite{chen2024improve, qiao2025mesh4d}, we quantified temporal smoothness by the mean of the second-order finite differences of the BiV volume trajectory, denoted as volume smoothness, and vertex trajectories, denoted as mesh smoothness. Cycle closure errors were evaluated by volume cycle consistency and mesh cycle consistency, which represent start-to-end discrepancies. To assess clinical relevance, we derived standard ventricular functional indices from the full-cycle left ventricle (LV) and right ventricle (RV) volume trajectories, including end-diastolic volume (EDV), end-systolic volume (ESV), ejection fraction (EF), which are routinely reported in cardiovascular magnetic resonance (CMR) examinations \cite{hundley2022society,puntmann2018society}. We computed the mean absolute error (MAE) of these functional indices between the ground truth sequence and the generated sample paired by subject covariates. 

\subsection{Experimental Results}
\noindent\textbf{Quantitative Analysis of the Generation Quality:}
As shown in Table~\ref{tab:ed_metrics}, we conducted unconditional and conditional cardiac cycle generation to comprehensively demonstrate the superiority of our method. In the unconditional setting, 4D F-MeshLDM$_{\text{uncond}}$ achieved the best results across all metrics and anatomies. For BiV, Coverage@5 achieved 71.2\%, while specificity improved from $2.41\pm0.53$ mm to $2.32\pm0.36$ mm. The largest fidelity gain was observed for RV specificity, which decreased from $2.89\pm0.56$ mm to $2.41\pm0.34$ mm. In the conditional generation, 4D F-MeshLDM$_{\text{cond}}$ achieved the best specificity and Coverage@5 for all anatomies. For BiV, Coverage@5 increased to 98.60\%, compared with 92.79\% for CHeart. 4D F-MeshLDM$_{\text{cond}}$ also achieved the best BiV specificity with $2.76\pm0.52$ mm, improving over CHeart ($2.80\pm0.51$ mm) and CardiacFlow ($2.95\pm0.56$ mm). 4D F-MeshLDM$_{\text{cond}}$ consistently achieved the lowest sequence reconstruction errors across all anatomical structures. Specifically, for BiV, our method reduced the Seq. RMSE to $7.54\pm2.75$ mm, substantially outperforming both CHeart ($8.19\pm2.97$ mm) and CardiacFlow ($9.07\pm3.32$ mm). These results demonstrate that our framework not only preserves the diversity and fidelity of the population distribution but also more accurately captures the precise, subject-specific dynamic motion trajectories of individual patients.\\

\noindent\textbf{Temporal Consistency:}
To evaluate the cycle consistency, we adapted 4D CardioSynth to conditional model and generated conditional BiV mesh sequences for each method. Table~\ref{tab:cycle_metrics} shows that 4D F-MeshLDM provides substantially stronger temporal cycle consistency than prior methods. Volume cycle consistency was reduced to $4.506\times 10^{-6}$ mL$^2$ and mesh cycle consistency was reduced to (5.564$\pm$0.309)$\times$10$^{-3}$ mm. Because of the Fourier parameterisation, our 4D F-MeshLDM actually yields a theoretical zero cycle closure error. The non-zero values are solely due to the normal cardiac cycle drift from the data, which is nearly-zero for populations without severe pathologies, whereas baselines suffer from structural drift. A lower second-order difference values of smoothness indicate smoother temporal evolution, but excessive smoothness may not be consistent with the true systolic-diastolic dynamics of the cardiac cycle. 4D F-MeshLDM achieves the lowest mesh smoothness and near-zero cycle closure error, indicating temporally coherent motion without endpoint drift. Overall, the near-zero cycle closure errors in both volume and mesh metrics and the best mesh smoothness strongly support the effectiveness of the periodic latent trajectory. Regarding inference time, marginally slower than CHeart (0.0919s), CardiacFlow (0.0941s), and 4D CardioSynth (0.0945s), our 4D F-MeshLDM still achieves rapid generation, producing an entire 3D+t sequence in just 0.1630s.\\

\noindent\textbf{Clinical relevance:}
Table~\ref{tab:cycle_metrics} also demonstrates the results of clinical relevance evaluation. Since $\mathrm{EF}=1-\mathrm{ESV}/\mathrm{EDV}$ which shows that correlated EDV and ESV errors can partially cancel in the ratio, the lower EF MAE indicates better preservation of relative pump function of 4D F-MeshLDM, while the absolute EDV/ESV errors of the proposed model remain comparable to the baselines rather than uniformly superior.\\

\begin{table}[t]
\centering
\caption{Quantitative Analysis of cardiac cycle generation. Specificity and Seq. RMSE are shown as mean$\pm$std.}
\label{tab:ed_metrics}

\scriptsize 

\begin{adjustbox}{max width=\textwidth}
\begin{tabular}{@{}l|ccc|ccc|ccc@{}} 
\toprule
\multirow{2}{*}{Method} 
& \multicolumn{3}{c|}{Specificity (mm, $\downarrow$)} 
& \multicolumn{3}{c|}{Coverage@5 (\%, $\uparrow$)}
& \multicolumn{3}{c}{Seq. RMSE (mm, $\downarrow$)} \\
\cmidrule(lr){2-4}\cmidrule(lr){5-7}\cmidrule(l){8-10}
& LV & RV & BiV & LV & RV & BiV & LV & RV & BiV \\
\midrule

\multicolumn{10}{c}{\textit{Unconditional Generation}} \\
\midrule

4D CardioSynth
& 1.94$\pm$0.33 & 2.89$\pm$0.56 & 2.41$\pm$0.53
& 60.2 & 66.5 & 64.4
& - & - & - \\

4D F-MeshLDM$_{\text{uncond}}$
& \textbf{1.89$\pm$0.32} & \textbf{2.41$\pm$0.37} & \textbf{2.32$\pm$0.36}
& \textbf{63.5} & \textbf{72.8} & \textbf{71.2}
& - & - & - \\

\midrule

\multicolumn{10}{c}{\textit{Conditional Generation}} \\
\midrule

CHeart
& 2.32$\pm$0.49 & 2.92$\pm$0.56 & 2.80$\pm$0.51
& 91.40 & 93.02 & 92.79
& 7.67$\pm$2.79 & 8.63$\pm$3.23 & 8.19$\pm$2.97 \\

CardiacFlow
& 2.40$\pm$0.56 & 3.08$\pm$0.59 & 2.95$\pm$0.56
& 96.05 & 96.44 & 96.98
& 8.51$\pm$3.13 & 9.58$\pm$3.59 & 9.07$\pm$3.32 \\

4D F-MeshLDM$_{\text{cond}}$
& \textbf{2.27$\pm$0.51} & \textbf{2.89$\pm$0.34} & \textbf{2.76$\pm$0.52}
& \textbf{98.14} & \textbf{99.07} & \textbf{98.60}
& \textbf{6.82$\pm$2.41} & \textbf{7.97$\pm$2.91} & \textbf{7.54$\pm$2.75} \\

\bottomrule
\end{tabular}
\end{adjustbox}
\end{table}

\begin{table}[t]
\centering
\caption{Quantitative analysis of temporal consistency and functional indices. (Mesh metrics: mean$\pm$std. Volume/Mesh cycle consistency: MSE/RMSE between first and last frame meshes.)}
\label{tab:cycle_metrics}

\scriptsize

\begin{adjustbox}{max width=\textwidth}
\begin{tabular}{@{}c|cccc|ccc|ccc@{}}
\toprule

 & \multicolumn{4}{c|}{Temporal metrics ($\downarrow$)}
 & \multicolumn{3}{c|}{LV MAE ($\downarrow$)}
 & \multicolumn{3}{c}{RV MAE ($\downarrow$)} \\
\cmidrule{2-5}\cmidrule{6-8}\cmidrule{9-11}
Method
& \shortstack{Vol.\\smooth.}
& \shortstack{Mesh\\smooth.}
& \shortstack{Vol. Consistency \\ (mL$^2$)}
& \shortstack{Mesh Consistency\\(mm)}
& \shortstack{EDV\\(mL)}
& \shortstack{ESV\\(mL)}
& EF
& \shortstack{EDV\\(mL)}
& \shortstack{ESV\\(mL)}
& EF \\
\midrule

CHeart
& 0.548 & 0.058$\pm$0.026 & 14.496 & 1.106$\pm$0.396
& 15.21 & \textbf{13.10} & 0.062
& 9.21 & 5.10 & 0.058 \\

CardiacFlow
& 0.229 & 0.046$\pm$0.010 & 0.133 & 0.143$\pm$0.019
& 23.09 & 21.23 & 0.073
& 12.57 & 7.04 & 0.062 \\

4D CardioSynth
& \textbf{0.140} & 0.037$\pm$0.006 & 4.352 & 0.861$\pm$0.216
& \textbf{14.25} & 13.17 & 0.080
& \textbf{8.72} & \textbf{4.95} & 0.055 \\

4D F-MeshLDM
& 0.361 & \textbf{0.029$\pm$0.017} & \textbf{4.506$\times$10$^{-6}$} & \textbf{(5.564$\pm$0.309)$\times$10$^{-3}$}
& 18.86 & 16.00 & \textbf{0.049} 
& 9.91 & 5.76 & \textbf{0.047} \\

\bottomrule
\end{tabular}
\end{adjustbox}
\end{table}

\begin{figure}[t]
    \centering
    \includegraphics[width=\textwidth]{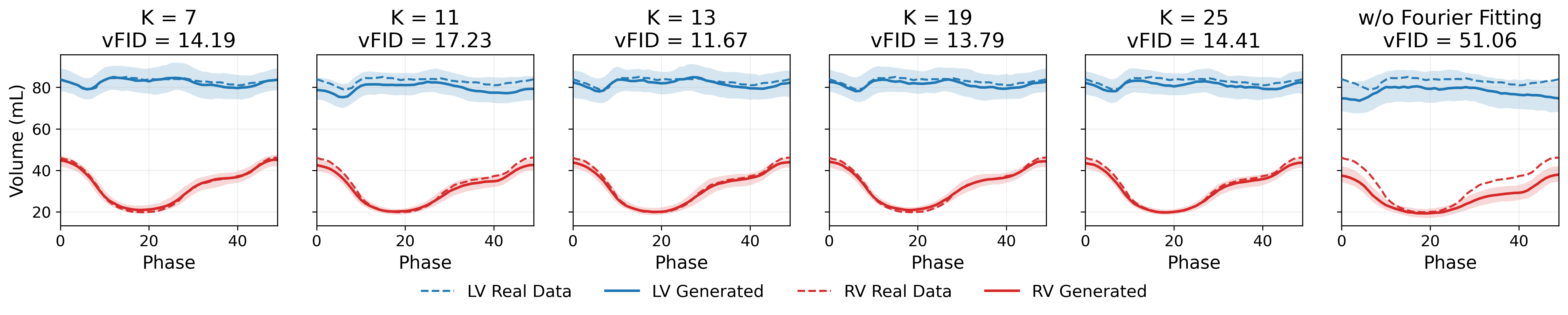}
    \caption{Ablation study on the Fourier latent trajectory model with the different truncation frequency $K$ on LV/RV volume trajectories.}
    \label{fig:ablation}
\end{figure}

\noindent\textbf{Ablation Study:} We conducted the ablation study on the impact of the Fourier parameterisation by varying the frequency $K$. The vFID was computed as a Fr\'echet distance between the distributions of real and generated volume trajectories \cite{ma2025cardiacflow}. Fig.~\ref{fig:ablation} visualises the mean$\pm$std volume curves for generated sequences against real data. Removing Fourier fitting substantially degraded trajectory realism with vFID $=51.06$, and the generated data exhibited significant bias in both LV and RV volumes over the cycle. The Fourier fitting consistently improved vFID and aligned the systolic-diastolic dynamics with the real distribution. The best performance was achieved at $K=13$ with the lowest vFID. This result indicates that Fourier fitting plays a key role in realistic cardiac cycle generation.\\

\begin{figure}[t]
    \centering
    \includegraphics[width=0.9\textwidth]{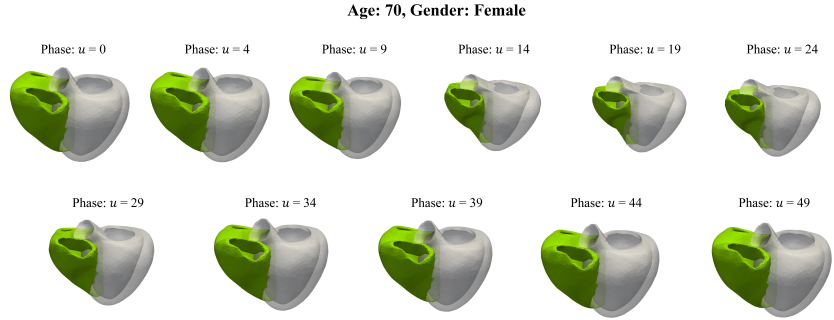}
    \caption{Visual qualitative example of a generated cardiac cycle from 4D F-MeshLDM.}
    \label{fig:visual}
\end{figure}

\noindent\textbf{Visual Assessment:}
Fig.~\ref{fig:visual} visualises a BiV mesh sequence generated by 4D F-MeshLDM conditioned on age 70 and female sex, which exhibits a coherent contraction-relaxation pattern across the cycle. The surfaces preserve anatomically plausible global morphology throughout the motion and the sequence demonstrates cycle consistency.

\section{Conclusion}
This paper introduced 4D F-MeshLDM, a conditional latent diffusion model for full-cycle 3D+t cardiac mesh synthesis with exact periodicity. A mesh VAE encodes phase anatomy, while a truncated Fourier representation enforces cycle-consistent latent trajectories whose coefficient tokens are modelled by a conditional Transformer DDPM. On 5{,}000 UK Biobank subjects, our method demonstrated high generation fidelity and achieved near-zero cycle closure error. Future work will investigate the performance of this work on populations with diverse cardiac cycle durations. We will also explore relaxing the fixed-topology limitation to better capture severe pathologies. Conditioning will be further extended to more diverse clinical covariates for downstream digital-twin and in-silico trials pipelines.

\bibliographystyle{splncs04}
\bibliography{main}

\end{document}